\documentclass[a4paper]{article}
\usepackage[margin=1.5in]{geometry}
\usepackage{makeidx}
\usepackage{graphicx}
\usepackage{amsmath,amssymb} 
\usepackage{color}

\usepackage[pagebackref=true,breaklinks=true,colorlinks,bookmarks=false]{hyperref}

\usepackage{url}
\urldef{\mailsa}\path|{boettger, follmann, fauser}@mvtec.com|

\usepackage[caption=false]{subfig}

\usepackage{xspace}
\newcommand{\latinphrase}[1]{\textit{#1}}  
\newcommand{\etal}{\latinphrase{et~al.}\xspace}

\newcommand{\figref}[1]{\mbox{Fig.~\ref{#1}}}
\newcommand{\tabref}[1]{\mbox{Table~\ref{#1}}}

\usepackage{pgfplots}
\usepackage{pgf,tikz}
\usepackage{mathrsfs}
\usetikzlibrary{calc}
\usetikzlibrary{arrows}
\pgfplotsset{compat=1.14}

\definecolor{colorNoScale}{RGB}{0,0,128}
\definecolor{colorNoRot}{RGB}{0,14,255}
\definecolor{colorRect2}{RGB}{0,254,222}
\definecolor{colorKCF}{RGB}{45,236,0}
\definecolor{colorDSST}{RGB}{159,235,36}
\definecolor{colorCCOT}{RGB}{255,228,0}
\definecolor{colorANT}{RGB}{255,96,4}
\definecolor{colorL1APG}{RGB}{255,0,230}
\definecolor{colorLGT}{RGB}{236,132,239}

\definecolor{colorRectangle1}{RGB}{0,255,0}
\definecolor{colorRectangle2}{RGB}{0,12,255}
\definecolor{colorInnerCircle}{RGB}{255,0,255}
\definecolor{colorInnerRectangle}{RGB}{16,16,16}
\definecolor{colorRectangularity}{RGB}{255,196,0}

\newenvironment{customlegend}[1][]{%
    \begingroup
    \csname pgfplots@init@cleared@structures\endcsname
    \pgfplotsset{#1}%
}{%
    \csname pgfplots@createlegend\endcsname
    \endgroup
}%

\def\addlegendimage{\csname pgfplots@addlegendimage\endcsname}
\def\addlegendentry{\csname pgfplots@addlegendentry\endcsname}
\begin{document}

\pagestyle{headings}

\author{Tobias B\"ottger${}^{+*}$%
\and Patrick Follmann${}^{+*}$ \and Michael Fauser${}^{+}$}
\title{Measuring the Accuracy of Object Detectors and Trackers}
\date{${}^{+}$MVTec Software GmbH, Munich, Germany\\
${}^{*}$Technical University of Munich (TUM) \\
\mailsa\\
\today}
\maketitle

%
\begin{abstract}
	The accuracy of object detectors and trackers is most commonly evaluated by the Intersection over Union (IoU) criterion. To date, most approaches are restricted to axis-aligned or oriented boxes and, as a consequence, many datasets are only labeled with boxes. Nevertheless, axis-aligned or oriented boxes cannot accurately capture an object's shape. To address this, a number of densely segmented datasets has started to emerge in both the object detection and the object tracking communities. However, evaluating the accuracy of object detectors and trackers that are restricted to boxes on densely segmented data is not straightforward. To close this gap, we introduce the relative Intersection over Union (rIoU) accuracy measure. The measure normalizes the IoU with the optimal box for the segmentation to generate an accuracy measure that ranges between 0 and 1 and allows a more precise measurement of accuracies. Furthermore, it enables an efficient and easy way to understand scenes and the strengths and weaknesses of an object detection or tracking approach. We display how the new measure can be efficiently calculated and present an easy-to-use evaluation framework. The framework is tested on the DAVIS and the VOT2016 segmentations and has been made available to the community.
\end{abstract}

\section{Introduction}
\label{sec:introduction}
Visual object detection and tracking are two rapidly evolving research areas with dozens of new algorithms being published each year. To compare the performance of the many different approaches, a vast amount of evaluation datasets and schemes are available. They include large detection datasets with multiple object categories, such as PASCAL VOC \cite{everingham_2015_pascal_voc}, smaller, more specific detection datasets with a single category, such as cars \cite{roman_2015_cod20k}, and sequences with multiple frames that are commonly used to evaluate trackers such as VOT2016 \cite{vot_2016}, OTB-2015 \cite{wu_otb_2015}, or MOT16 \cite{mot_2016}. Although very different in their nature, all of the benchmarks use axis-aligned or oriented boxes as ground truth and estimate the accuracy with the Intersection over Union (IoU) criterion. 
	
\begin{figure}[!t]
\centering
  \subfloat[{\tt bag} from VOT2016 \cite{vot_2016}] {\includegraphics[width=0.32\textwidth]{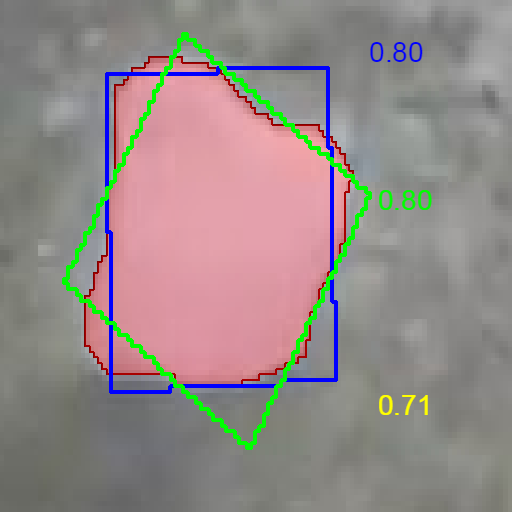}}
  \hfill	
\subfloat[{\tt blackswan} from DAVIS \cite{perazzi_2016_benchmark}] {\includegraphics[width=0.32\textwidth]{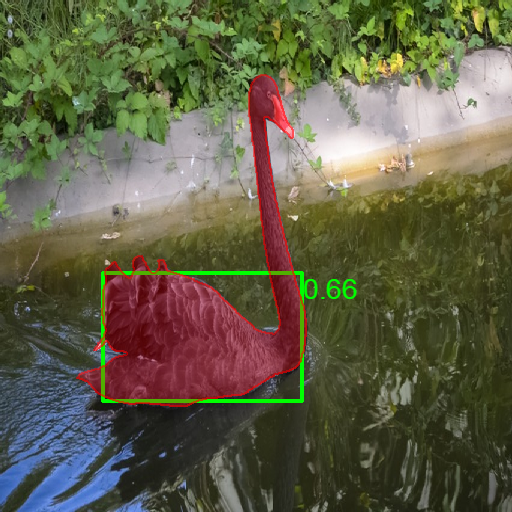}}
  \hfill  
  \subfloat[{\tt boat} from DAVIS \cite{perazzi_2016_benchmark}] {\includegraphics[width=0.32\textwidth]{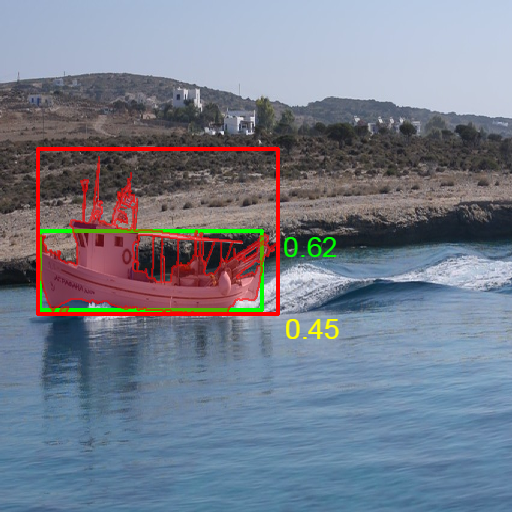}}
  \hfill
  
\caption{In image (a), both oriented boxes have an identical IoU with the ground truth segmentation. Nevertheless, their common IoU is only 0.71. Restricting the ground truth to boxes may introduce an undesired bias in the evaluation. In image (b), the best possible IoU of an axis-aligned box is only 0.66. Hence, for segmented data, it is difficult to use the absolute value of the IoU as an accuracy measure since it generally does not range from $0$ to $1$. Furthermore, although the object detection (green) in image (c) has an overlap of 0.62 with the ground truth segmentation, its IoU with the ground truth axis-aligned bounding box is only 0.45 and would be considered a false detection in the standard procedure. The proposed rIoU is the same for both boxes in (a) and 1.0 for the green boxes in (b) and (c).}
\label{fig:rectangle1example}
\end{figure}

Nevertheless, boxes are very crude approximations of many objects and may introduce an unwanted bias in the evaluation process, as is displayed in \figref{fig:rectangle1example}. Furthermore, approaches that are not restricted to oriented or axis-aligned boxes will not necessarily have higher accuracy scores in the benchmarks. To address these problems, a number of densely segmented ground truth datasets has started to emerge \cite{lin_2014_coco,perazzi_2016_benchmark,voji_2017_pixel_wise}.

Unfortunately, evaluating the accuracy of object detectors and trackers that are restricted to boxes on densely segmented data is not straightforward. For example, the VOT2016 Benchmark \cite{vot_2016} generates plausible oriented boxes from densely segmented objects and the COCO 2014 Detection challenge \cite{lin_2014_coco} uses axis-aligned bounding boxes of the segmentations to simplify the evaluation protocol. Hence, approaches may have a relatively low IoU with the ground truth, although their IoU with the actual object segmentation is the same (or even better) than that of the ground truth box (see \figref{fig:rectangle1example}(c)).

To enable a fair evaluation of algorithms restricted to axis-aligned or oriented boxes on densely segmented data we introduce the relative Intersection over Union accuracy (rIoU) measure. The rIoU uses the best possible axis-aligned or oriented box of the segmentation to normalize the IoU score. The normalized IoU ranges from 0 to 1 for an arbitrary segmentation and allows to determine the true accuracy of a scheme. For tracking scenarios, the optimal boxes have further advantages. By determining three different optimal boxes for each sequence, the optimal oriented box, the optimal axis-aligned box and, the optimal axis-aligned box for a fixed scale, it is possible to identify scale changes, rotations, and occlusion in a sequence without the need of by-frame labels. 

The optimal boxes are obtained in a fast and efficient optimization process. We validate the quality of the boxes in the experiments section by comparing them to a number of exhaustively determined best boxes for various scenes.

The three main contributions of this paper are:
\begin{enumerate}
\item The introduction of the relative Intersection over Union accuracy (rIoU) measure, which allows an accurate measurement of object detector and tracker accuracies on densely segmented data.
\item The proposed evaluation removes the bias introduced by restricting the ground truth to boxes for densely segmented data (such as COCO 2014 Detection Challenge \cite{lin_2014_coco} or VOT2016 \cite{vot_2016}).
\item A compact, easy-to-use, and efficient evaluation scheme for evaluating object trackers that allows a good interpretability of a trackers strengths and weaknesses.
\end{enumerate}
The proposed measure and evaluation scheme is evaluated on a handful of state-of-the-art trackers for the DAVIS \cite{perazzi_2016_benchmark} and VOT2016 \cite{vot_2016} datasets and will be made available to the community.

%
\section{Related Work}
\label{sec:relatedwork}

In the object detection community, the most commonly used accuracy measure is the 
Intersection over Union (IoU), also called Pascal overlap or bounding box overlap \cite{everingham_2015_pascal_voc}. It is commonly used as the standard requirement for a correct detection, when the IoU between the predicted detection and the ground truth is at least $0.5$ \cite{lin_2014_coco}.

In the tracking community, many different accuracy measures have been proposed, most of them center-based and overlap-based measures \cite{cehovin_2016_visual,vot_2016,Kristan_2016_novel,nawaz_2013_protocol,smeulders_2014_visual,wu_otb_2015}. To unify the evaluation of trackers, \v{C}ehovin \etal \cite{cehovin_2014_new_tracker,cehovin_2016_visual} provide a highly detailed theoretical and experimental analysis of the most popular performance measures and show that many of the accuracy measures are highly correlated. Nevertheless, the appealing property of the IoU measure is that it accounts for both position and size of the prediction and ground truth simultaneously. This has lead to the fact that, in recent years, it has been the most commonly used accuracy measure in the tracking community \cite{vot_2016,wu_otb_2015}. For example, the VOT2016 \cite{vot_2016} evaluation framework uses the IoU as the sole accuracy measure and identifies tracker failures when the IoU between the predicted detection and the ground truth is $0.0$ \cite{Kristan_2016_novel}.

Since bounding boxes are very crude approximations of objects \cite{lin_2014_coco} and cannot accurately capture an object's shape, location, or characteristics, numerous datasets with densely segmented ground truth have emerged. For example, the COCO 2014 dataset \cite{lin_2014_coco} includes more than 886,000 densely annotated instances of 80 categories of objects. Nevertheless, on the COCO detection challenge the segmentations are approximated by axis-aligned bounding boxes to simplify the evaluation. As stated earlier, this introduces an unwanted bias in the evaluation. A further dataset with excellent pixel accurate segmentations is the DAVIS dataset \cite{perazzi_2016_benchmark}, which was released in 2017. It consists of 50 short sequences of manually segmented objects which, although originally for video object segmentation, can also be used for the evaluation of object trackers. Furthermore, the segmentations used to generate the VOT2016 ground truths have very recently been released \cite{voji_2017_pixel_wise}.

In our work, we enable the evaluation of object detection and tracking algorithms that are restricted to output boxes on densely segmented ground truth data. The proposed approach is easy to add to existing evaluations and improves the precision of the standard IoU accuracy measure.

%
\section{Relative Intersection over Union (rIoU)}
Using segmentations for evaluating the accuracy of detectors or trackers removes the bias a bounding-box abstraction induces. Nevertheless, the IoU of a box and an arbitrary segmentation generally does not range from $0$ to $1$, where the maximum value depends strongly on the objects' shape.
For example, in \figref{fig:rectangle1example}(b) the best possible axis-aligned box only has an  IoU of 0.66 with the segmentation.

To enable a more precise measurement of the accuracy, we introduce the relative Intersection over Union (rIoU) of a box $\mathcal{B}$ and a dense segmentation $\mathcal{S}$ as
\begin{equation}
\label{eq:rIoU}
\Phi_{rIoU} \left( \mathcal{S}, \mathcal{B}\right ) = \frac{\Phi_{IoU}(\mathcal{S},\mathcal{B})}{\Phi_{opt}(\mathcal{S})},
\end{equation}
where $\Phi_{IoU}$ is the Intersection over Union (IoU),
\begin{equation}
\label{eq:IoU}
\Phi_{IoU} \left( \mathcal{S}, \mathcal{B}\right ) = \frac{\left \vert \mathcal{S} \cap \mathcal{B} \right \vert}{\left \vert \mathcal{S} \cup \mathcal{B} \right \vert},
\end{equation}
and $\Phi_{opt}$ is the best possible IoU a box can achieve for the segmentation $\mathcal{S}$.
In comparison to the usual IoU ($\Phi_{IoU}$), the rIoU measure ($\Phi_{rIoU}$) truly ranges from $0$ to $1$  for all possible segmentations. Furthermore, the measure makes it possible to interpret ground truth attributes such as scale change or occlusion, as is displayed later in section \ref{sec:theoretical}.

The calculation of $\Phi_{opt}$, required to obtain $\Phi_{rIoU}$, is described in the following section.

\subsection{Optimization}

An oriented box $\mathcal{B}$ can be parameterized with 5 parameters
\begin{equation}
\label{eq:boxparam}
b = \left (r_c,c_c,w,h,\phi\right ),
\end{equation}
where $r_c$ and $c_c$ denote the row and column of the center, $w$ and $h$ denote the width and height, and $\phi$ the orientation of the box with respect to the column-axis. An axis-aligned box can equally be parameterized with the above parameters by fixing the orientation to $0^\circ$.

\begin{figure}
\centering
 {\includegraphics[width=0.4\textwidth]{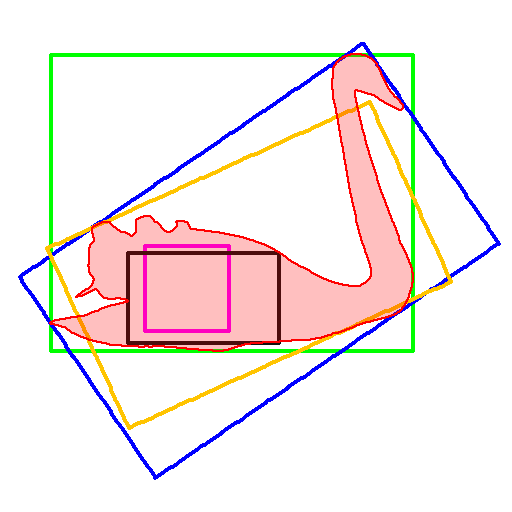}}
  \hfill
\caption{{\tt blackswan} from DAVIS \cite{perazzi_2016_benchmark}. The initial values of the optimization process of \eqref{eq:optimization} are displayed. We use the axis-aligned bounding box (green), the oriented bounding box (blue), the inner square of the largest inner circle (magenta), the largest inner axis-aligned box (black) and the oriented box with the same second order moments as the segmentation (orange).}
\label{fig:initialvalues}
\end{figure}

For a given segmentation $\mathcal{S}$, the box with the best possible IoU is
\begin{equation}
\label{eq:optimization}
\Phi_{opt}(\mathcal{S}) = \max_b\,\,\, \Phi_{IoU} (\mathcal{S},\mathcal{B}(b)) \hspace{1cm} s.t.\,\, b \in \mathbb{R}_{>0}^4 \times [0^\circ,90^\circ).
\end{equation}
For a convex segmentation, the above problem can efficiently be optimized with the method of steepest descent. To handle arbitrary, possibly unconnected, segmentations, we optimize \eqref{eq:optimization} with a multi-start gradient descent with a backtracking line search. The gradient is approximated numerically by the symmetric difference quotient.
We use the diverse set of initial values for the optimization process displayed in \figref{fig:initialvalues}. The largest axis-aligned inner box (black) and the inner box of the largest inner circle (magenta) are completely within the segmentation. Hence, in the optimization process, they will gradually grow and include background if it improves $\Phi_{IoU}$. On the other hand, the bounding boxes (green and blue) include the complete segmentation and will gradually shrink in the optimization to include less of the segmentation. The oriented box with the same second order moments as the segmentation (orange) serves as an intermediate starting point \cite{rosin_1999_measuring}. Hence, only if the initial values converge to different optima do we need to expend more effort. In these cases, we randomly sample further initial values from the interval spanned by the obtained optima with an added perturbation. In our experiements we used 50 random samples. Although this may lead to many different optimizations, the approach is still very efficient. A single evaluation of $\Phi_{IoU} (\mathcal{S},\mathcal{B})$ only requires an average of $0.04$ms for the segmentations within the DAVIS \cite{perazzi_2016_benchmark} dataset in HALCON\footnote{MVTec Software GmbH, \url{https://www.mvtec.com/}} on an IntelCore i7-4810 CPU @2.8GHz with 16GB of RAM with Windows 7 (x64). As a consequence, the optimization of $\Phi_{opt}$ requires an average of 1.3s for the DAVIS \cite{perazzi_2016_benchmark} and 0.7s for the VOT2016 \cite{vot_2016} segmentations.

\subsection{Validation}
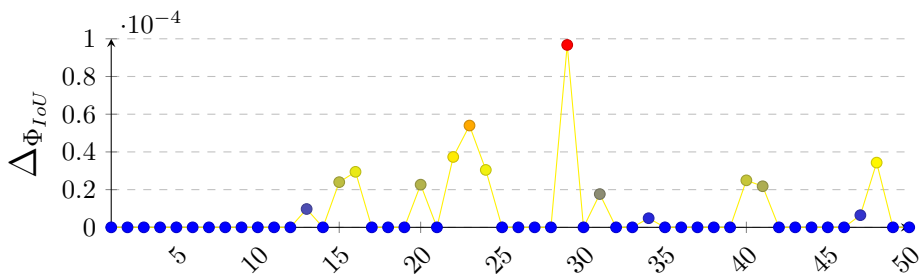
\begin{figure}
\centering
\begin{tikzpicture}
  \begin{axis}[
  	scatter,
	scale only axis=true,
	width=10.5cm, height=2.5cm,
	y label style={at={(axis description cs:-0.07,.5)},anchor=south},
    ylabel={\Large$\Delta_{\Phi_{IoU}}$},
    xmin=1, xmax=50,
    ymin=0, ymax=0.0001,
    ytick={0,0.00002,0.00004,0.00006,0.00008,0.0001},
    x tick label style={rotate=45, anchor=north east, inner sep=1mm},
    legend pos=north west,
    ymajorgrids=true,
    grid style=dashed,
    axis lines=left,
]
 
\addplot [color=yellow] table {validate_no_rot.dat};
\end{axis}
\end{tikzpicture}
\hfill
\caption{The absolute difference $\Delta_{\Phi_{IoU}}$ of the exhaustively determined best axis-aligned box and the optimized axis-aligned box for a selected frame in each of the 50 DAVIS \cite{perazzi_2016_benchmark} sequences. Most boxes are identical, only a handful of boxes are marginally different $(<0.0001$).}
\label{fig:validation}
\end{figure}

To validate the optimization process, we exhaustively searched for the best boxes in a collection of exemplary frames from each of the 50 sequences in the DAVIS dataset \cite{perazzi_2016_benchmark}. The validation set consists of frames that were challenging for the optimization process. In a first step, we validated the optimization for axis-aligned boxes. The results in \figref{fig:validation} indicate that the optimization is generally very close or identical to the exhaustively determined boxes.

For the oriented boxes, one of the restrictions we can make is that the area must at least be as large as the smallest inner box of the segmentation and may not be larger than the bounding oriented box. Nevertheless, even with further heuristics, the number of candidates to test is in the number of billions for the sequences in the DAVIS dataset. Given a pixel-precise discretization for $r_c,c_c,w,h$ and a $0.5^\circ$ discretization of $\phi$, it was impossible to find boxes with a better IoU than the optimized oriented boxes in the validation set. This is mostly due to the fact that the sub-pixel precision of the parameterization (especially in the angle $\phi$) is of paramount importance for the IoU of oriented boxes.

%

\section{Theoretical Trackers}
\label{sec:theoretical}
The concept of theoretical trackers was first introduced by \v{C}ehovin \etal \cite{cehovin_2016_visual} as an ``\textit{excellent interpretation guide in the graphical representation of results}''. In their paper, they use perfectly robust or accurate theoretical trackers to create bounds for the comparison of the performance of different trackers. In our case, we use the boxes with an optimal IoU to create upper bounds for the accuracy of trackers that underlie the box-world assumption. We introduce three theoretical trackers that are obtained by optimizing \eqref{eq:optimization} for a complete sequence. Given the segmentation $\mathcal{S}$, the first tracker returns the best possible axis-aligned box ({\tt box-axis-aligned}), the second tracker returns the optimal oriented box ({\tt box-rot}) and the third tracker returns the optimal axis-aligned box with a fixed scale ({\tt box-no-scale}). The scale is initialized in the first frame with the scale of the box determined by {\tt box-axis-aligned}.

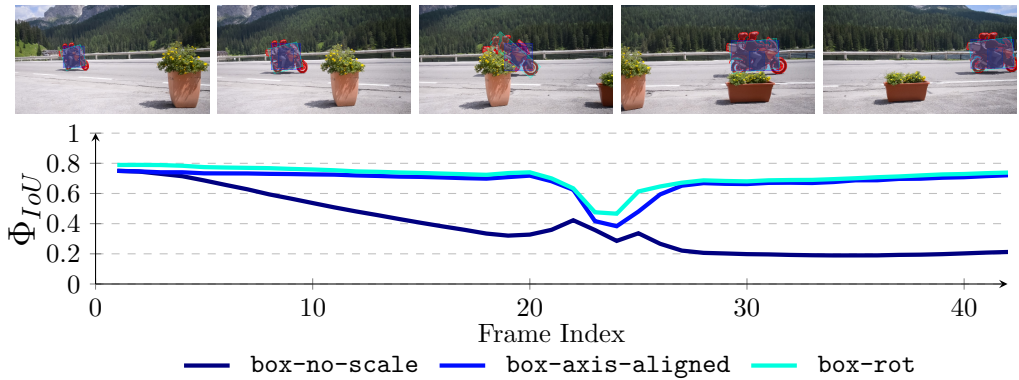
\begin{figure}[t]
{\flushright
\hspace*{0.1cm}
\includegraphics[width=0.19\textwidth]{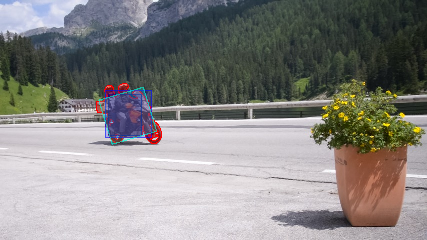}
\includegraphics[width=0.19\textwidth]{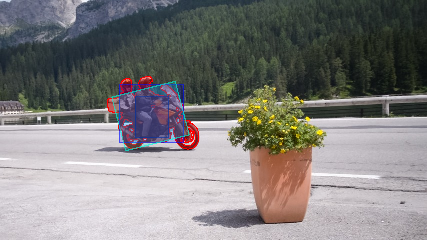}
\includegraphics[width=0.19\textwidth]{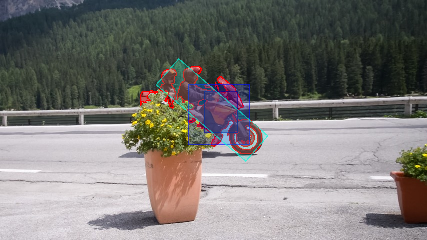}
\includegraphics[width=0.19\textwidth]{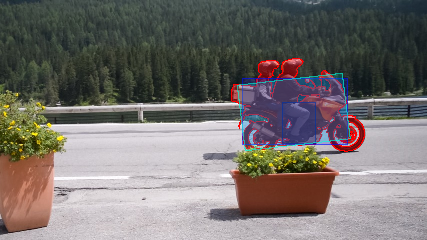}
\includegraphics[width=0.19\textwidth]{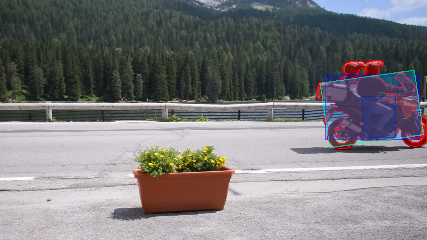}
}
\centering
\begin{tikzpicture}
\begin{axis}[
	scale only axis=true,
	width=12.cm, height=2.0cm,
	x label style={at={(axis description cs:0.5,-0.2)},anchor=north},
	y label style={at={(axis description cs:-0.05,.5)},anchor=south},
    xlabel={Frame Index},
    ylabel={\Large$\Phi_{IoU}$},
    xmin=0, xmax=42,
    ymin=0, ymax=1.00,
    xtick={0,10,20,30,40},
    ytick={0,.20,.40,.60,.80,1.00},
    legend style={at={(0.5, -0.4)}, anchor = north, column sep=5pt, draw=none},
    legend columns=3,
    ymajorgrids=true,
    grid style=dashed,
    axis lines=left,
    every axis plot/.append style={ultra thick}
]
 
\addplot [color=colorNoScale] table {theoretical_no_scale.dat};
\addplot [color=colorNoRot] table {theoretical_no_rot.dat};
\addplot [color=colorRect2] table {theoretical_rect2.dat};
\legend{{\tt box-no-scale}, {\tt box-axis-aligned}, {\tt box-rot}}
\end{axis}
\end{tikzpicture}
\hfill
\caption{{\tt motorbike} from DAVIS \cite{perazzi_2016_benchmark}. The increasing gap between the {\tt box-no-scale} and the other two theoretical trackers indicates a scale change of the motorbike. The drop in all three theoretical trackers around frame 25 indicates that the object is being occluded. The best possible IoU is never above 0.80 for the complete sequence.}
\label{fig:theoretical_tracker}
\end{figure}

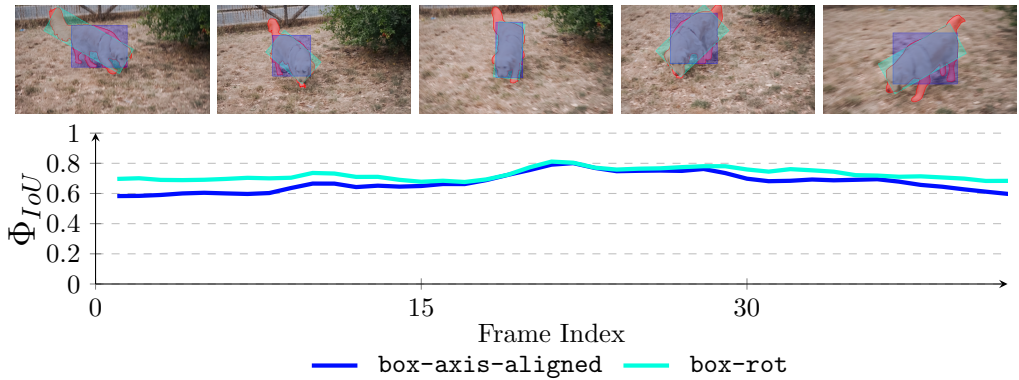
\begin{figure}[t]
{\flushright
\hspace*{0.1cm}
\includegraphics[width=0.19\textwidth]{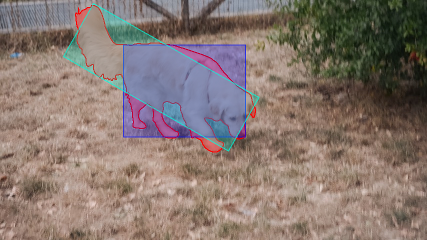}
\includegraphics[width=0.19\textwidth]{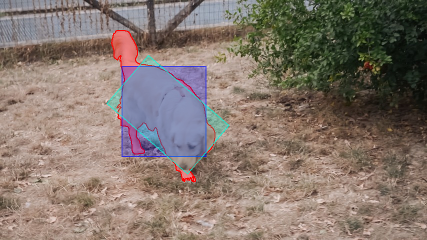}
\includegraphics[width=0.19\textwidth]{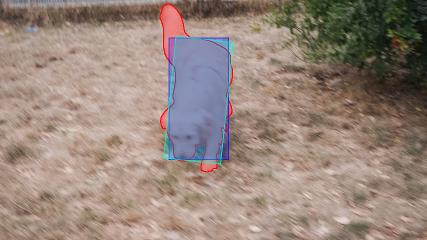}
\includegraphics[width=0.19\textwidth]{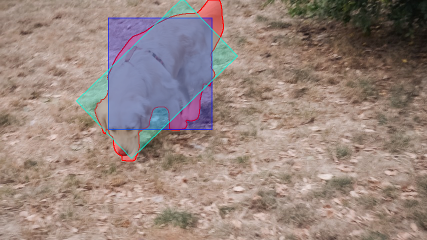}
\includegraphics[width=0.19\textwidth]{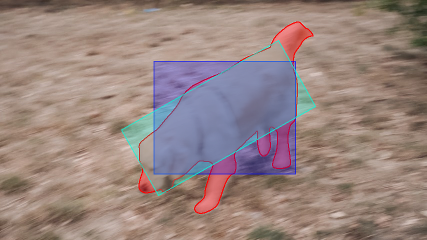}
}
\centering
\begin{tikzpicture}
\begin{axis}[
	scale only axis=true,
	width=12cm, height=2.0cm,
	x label style={at={(axis description cs:0.5,-0.2)},anchor=north},
	y label style={at={(axis description cs:-0.05,.5)},anchor=south},
    xlabel={Frame Index},
    ylabel={\Large$\Phi_{IoU}$},
    xmin=0, xmax=42,
    ymin=0, ymax=1.00,
    xtick={0,15,30,45,60},
    ytick={0,.20,.40,.60,.80,1.00},
    legend style={at={(0.5, -0.4)}, anchor = north, column sep=5pt, draw=none},
    legend columns=3,
    ymajorgrids=true,
    grid style=dashed,
    axis lines=left,
    every axis plot/.append style={ultra thick}
]
 
\addplot [color=colorNoRot] table {theoretical_no_rot_dog.dat};
\addplot [color=colorRect2] table {theoretical_rect2_dog.dat};
\legend{{\tt box-axis-aligned}, {\tt box-rot}}
\end{axis}
\end{tikzpicture}
\hfill
\caption{{\tt dog} from DAVIS \cite{perazzi_2016_benchmark}. The gaps between the {\tt box-axis-aligned} and {\tt box-rot} tracker indicate a rotation of the otherwise relatively compact segmentation of the dog. The best possible IoU is never above 0.80 for the complete sequence.}
\label{fig:theoretical_tracker_dog}
\end{figure}

The theoretical tracker can be used to normalize a tracker's IoU for a complete sequence, which enables a fair interpretation of a tracker's accuracy and removes the bias from the box-world assumption. 
Furthermore, the three different theoretical trackers make it possible to interpret a tracking scene without the need of by-frame labels. As is displayed in \figref{fig:theoretical_tracker}, the difference between the {\tt box-no-scale}, {\tt box-axis-aligned}, and {\tt box-rot} trackers indicates that the object is undergoing a scale change. Furthermore, the decreasing IoUs of all theoretical trackers indicate that the object is either being occluded or deforming to a shape that can be approximated less well by a box. For compact objects, the difference of the {\tt box-rot} tracker and the {\tt box-axis-aligned} tracker indicates a rotation or change of perspective, as displayed in \figref{fig:theoretical_tracker_dog}.

%
\section{Experiments}
We evaluate the accuracy of a handful of state-of-the art trackers on the DAVIS \cite{perazzi_2016_benchmark} and VOT2016 \cite{vot_2016} datasets with the new rIoU measure. We initialize the trackers with the best possible axis-aligned box for the given segmentation. Since we are primarily interested in the accuracy and not in the trackers robustness, we do not reinitialize the trackers when they move off target. Please note that the accuracy of the robustness measure is also improved when using segmentations; The failure cases (hence $\Phi_{IoU} = 0$) are identified earlier since $\Phi_{IoU}$ is zero when the tracker has no overlap with the segmentation and not with a bounding box abstraction of the object (which may contain a large amount of background, see, e.g., \figref{fig:rectangle1example}).

We restrict our evaluation to the handful of (open source) state-of-the-art trackers displayed in \tabref{tbl:compare_abs_rel_iou_davis}. A thorough evaluation and comparison of all top ranking trackers is beyond the scope of this paper. The evaluation framework is made available and constructed such that it is easy to add new trackers from MATLAB\footnote{The MathWorks, Inc., \url{https://www.mathworks.com/}}, Python\footnote{Python Software Foundation, \url{https://www.python.org/}} or HALCON. 

\begin{table}
\centering
\bgroup
\def\arraystretch{1.2} 
\setlength{\tabcolsep}{12pt}
\caption{Comparison of different tracking approaches and their average absolute ($\Phi_{IoU}$) and relative IoU ($\Phi_{rIoU}$) for the DAVIS \cite{perazzi_2016_benchmark} and the VOT2016 \cite{vot_2016} segmentations}
\label{tbl:compare_abs_rel_iou_davis}
\begin{tabular}{l c c  c c}
& \multicolumn{2}{c}{DAVIS } & \multicolumn{2}{c}{VOT2016} \\ \hline

  & $\Phi_{IoU}$ &  $\Phi_{rIoU}$ &  $\Phi_{IoU}$ &  $\Phi_{rIoU}$ \\ \hline
\multicolumn{5}{c}{Axis-aligned boxes (fixed scale)}\\ \hline
KCF \cite{henriques_2015_high_speed} & 0.40 & \textbf{0.78} & 0.23 & 0.45 \\ 
\multicolumn{5}{c}{Axis-aligned boxes} \\ \hline
DSST \cite{danelljan_2014_accuracte} & 0.43 & 0.67 & 0.24 & 0.32 \\
CCOT \cite{danelljan_2016_beyond} & \textbf{0.47} & 0.73 & \textbf{0.41} & \textbf{0.56} \\
ANT \cite{cehovin_2016_ant} & 0.40 & 0.64 & 0.26 & 0.37 \\
L1APG \cite{bao_2012_l1apg} & 0.40 & 0.63 & 0.18 & 0.25 \\
\multicolumn{5}{c}{Oriented boxes} \\ \hline
LGT \cite{cehovin_2013_tpami} & 0.40 & 0.60 & 0.25 & 0.34
\end{tabular}
\egroup
\end{table}

We include the Kernelized Correlation Filter (KCF) \cite{henriques_2015_high_speed} tracker since it was a top ranked tracker in the VOT2014 challenge even though it assumes the scale of the object to stay constant. The Discriminative Scale Space Tracker (DSST) \cite{danelljan_2014_accuracte} tracker is essentially an extension of KCF that can handle scale changes and outperformed the KCF by a small margin in the VOT2014 challenge. As further axis-aligned trackers, we include ANT \cite{cehovin_2016_ant}, L1APG \cite{bao_2012_l1apg}, and the best performing tracker from the VOT2016 challenge, the continuous convolution filters (CCOT) from Danelljan \emph{et al.} \cite{danelljan_2016_beyond}. We include the LGT \cite{cehovin_2013_tpami} as one of the few open source trackers that estimates the object position as an oriented box.

\begin{figure} [t]
\centering

\includegraphics[width=0.19\textwidth]{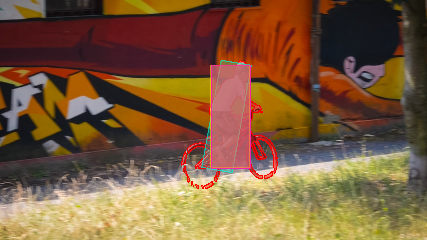}
\includegraphics[width=0.19\textwidth]{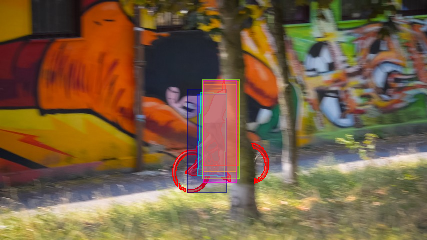}
\includegraphics[width=0.19\textwidth]{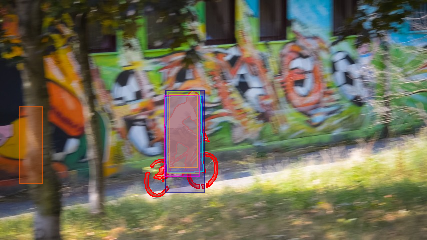}
\includegraphics[width=0.19\textwidth]{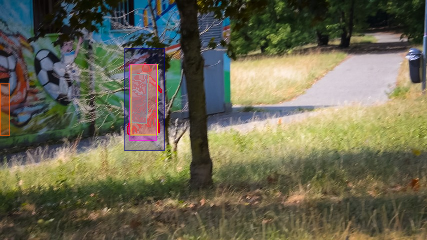}
\includegraphics[width=0.19\textwidth]{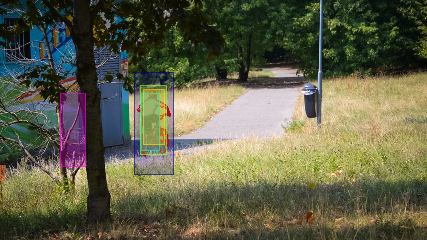}

\subfloat {\includegraphics[width=0.3\textwidth]{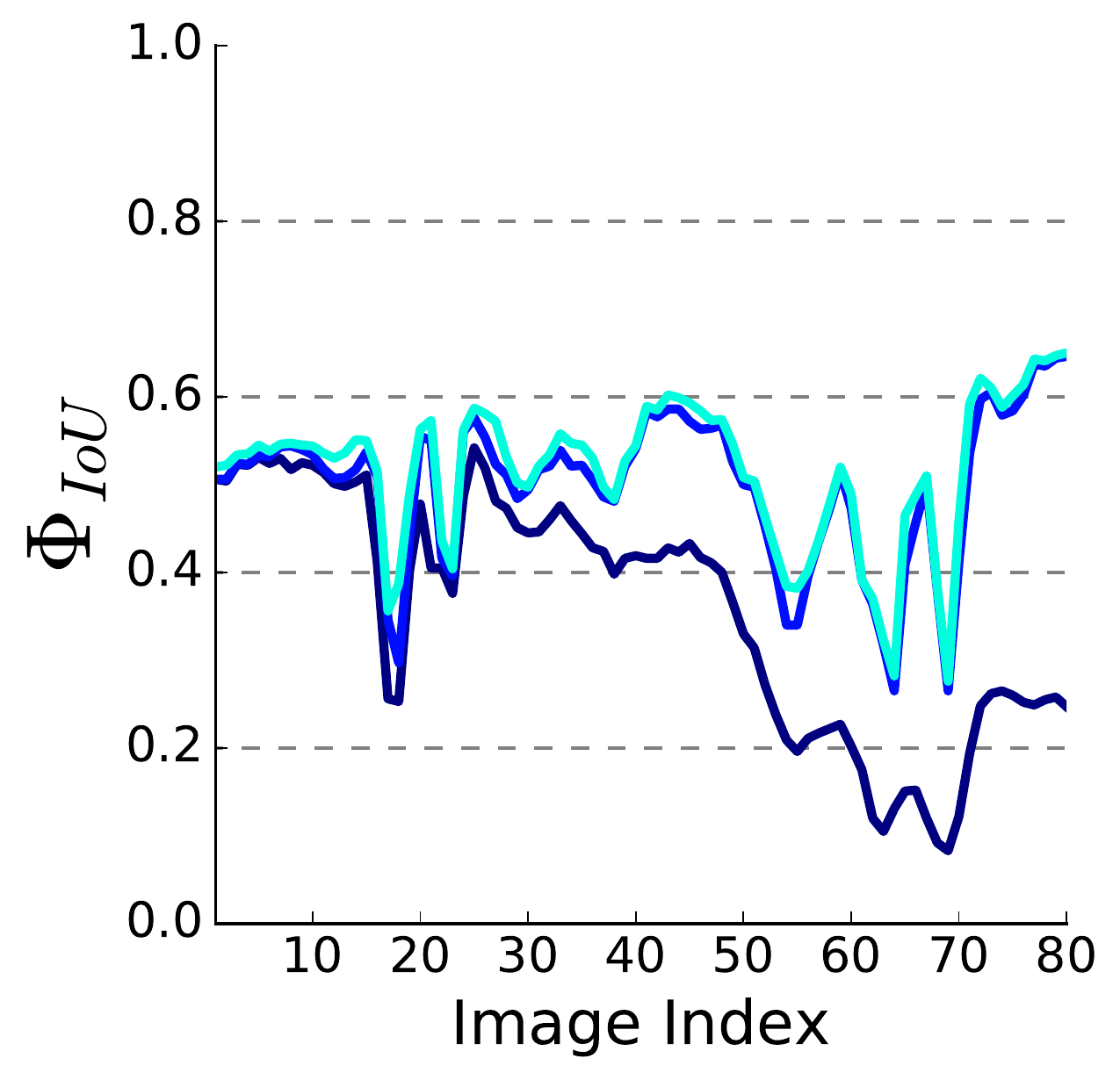}}
  \hfill
\subfloat {\includegraphics[width=0.3\textwidth]{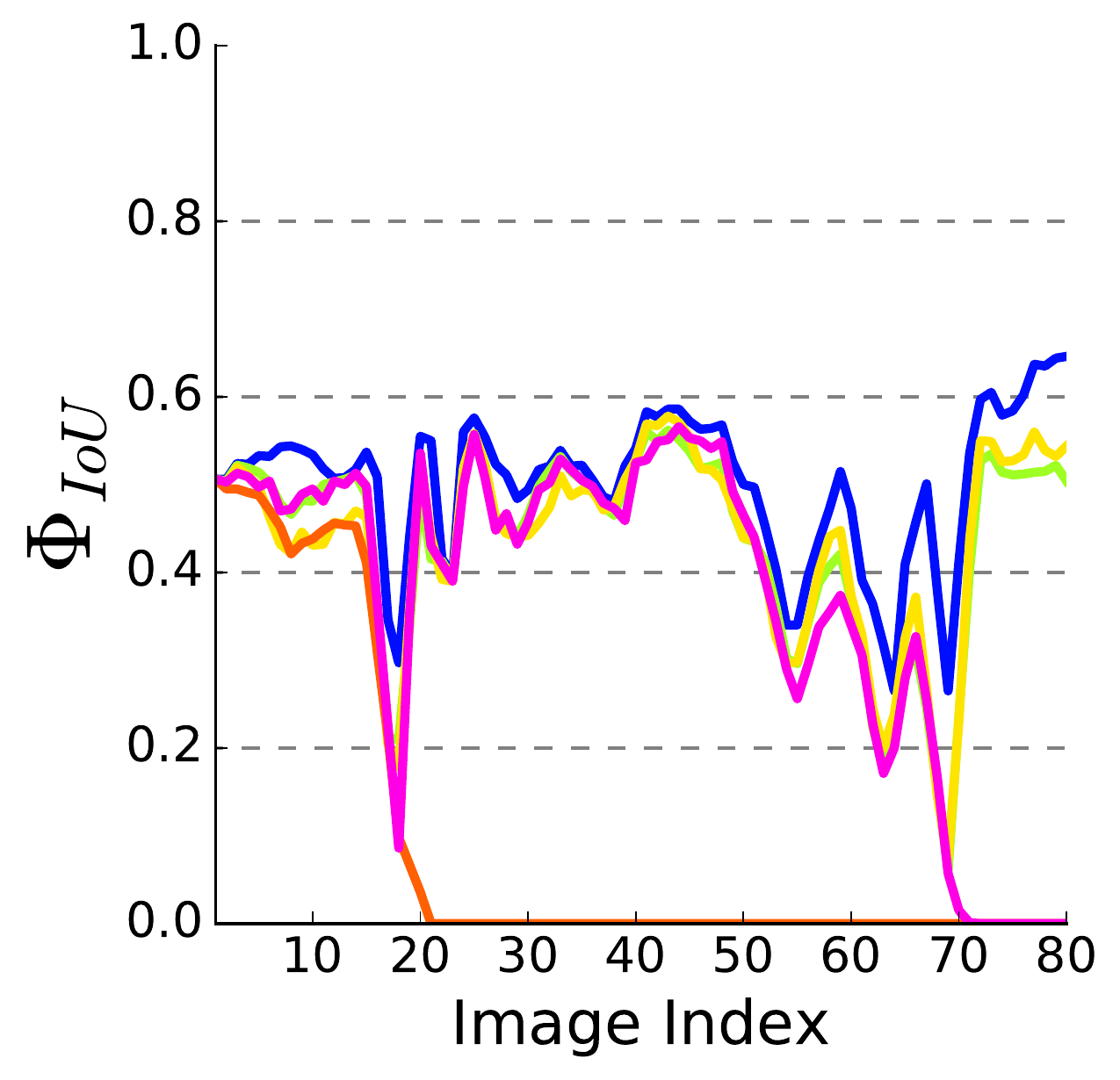}}
  \hfill
\subfloat {\includegraphics[width=0.3\textwidth]{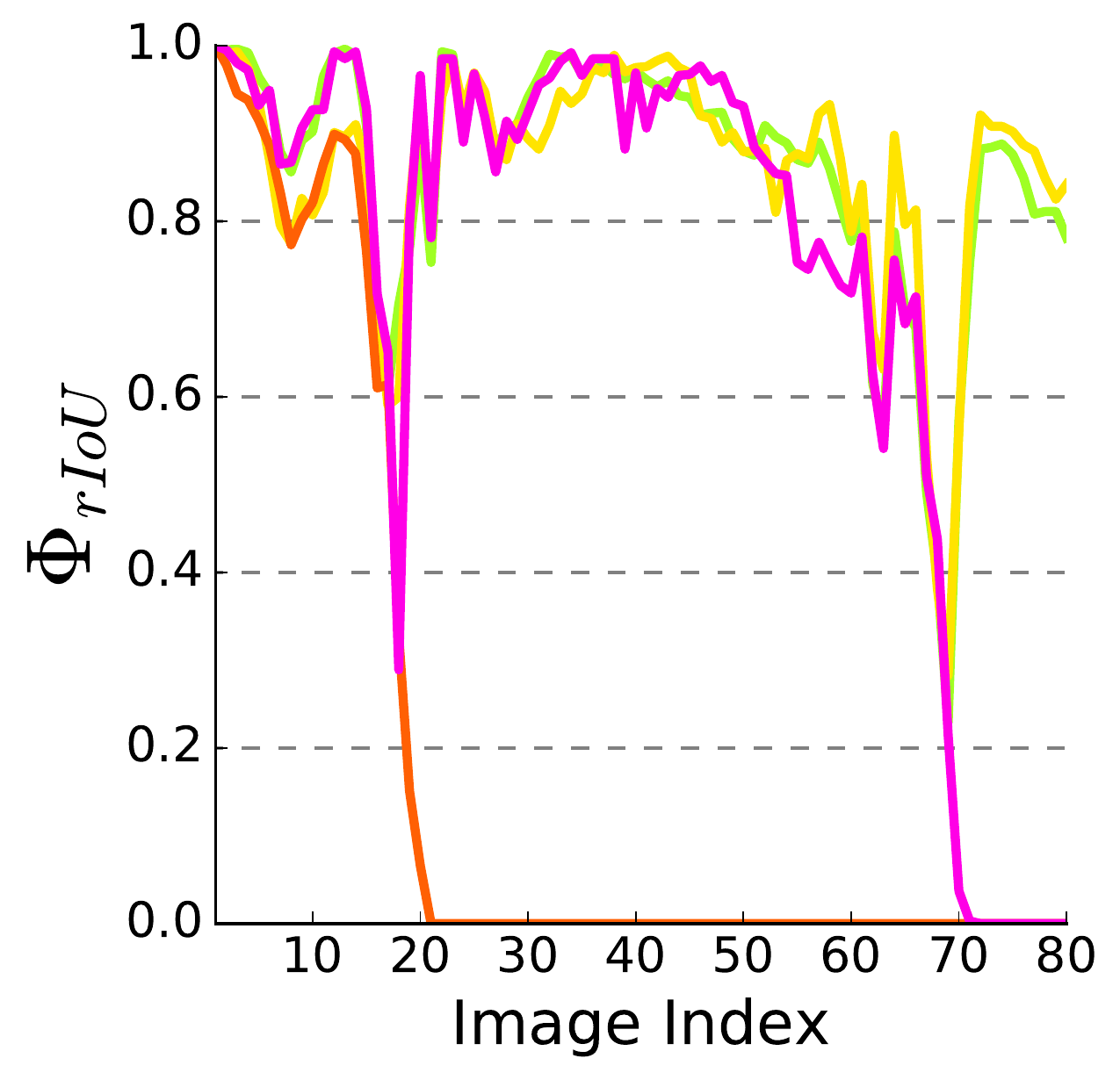}}
  \hfill
\begin{tikzpicture}
\begin{customlegend}[
   legend style={column sep=2pt, draw=none, line width=1.5pt}, 
   legend columns=4,
]
\addlegendentry{{\tt box-no-scale}}
\addlegendentry{{\tt box-axis-aligned}}
\addlegendentry{{\tt box-rot}}
\addlegendentry{}
\addlegendentry{DSST \cite{danelljan_2014_accuracte}}
\addlegendentry{CCOT \cite{danelljan_2016_beyond}}
\addlegendentry{ANT \cite{cehovin_2016_ant}}
\addlegendentry{L1APG \cite{bao_2012_l1apg}}

\addlegendimage{colorNoScale,fill=black!100!yellow,sharp plot}
\addlegendimage{colorNoRot,fill=black!100!yellow,sharp plot}
\addlegendimage{colorRect2,fill=black!100!yellow,sharp plot}
\addlegendimage{empty legend}
\addlegendimage{colorDSST,fill=black!100!yellow,sharp plot}
\addlegendimage{colorCCOT,fill=black!100!yellow,sharp plot}
\addlegendimage{colorANT,fill=black!100!yellow,sharp plot}
\addlegendimage{colorL1APG,fill=black!100!yellow,sharp plot}
\end{customlegend}
\end{tikzpicture}

\caption{{\tt bmx-trees} from DAVIS \cite{perazzi_2016_benchmark}. On the left, differences between {\tt box-no-scale} and {\tt box-axis-aligned} indicate that the object is changing scale and is occluded at frame 18 and around frames 60-70. In the middle plot, we compare the IoU of the axis-aligned box trackers and {\tt box-axis-aligned}. The corresponding rIoU plot is shown on the right. It becomes evident that the ANT tracker fails when the object is occluded for the first time and the L1APG tracker at the second occlusion. The rIoU shows that DSST and CCOT perform very well, while the IoU would imply they are relatively weak.}
\label{fig:experiments}
\end{figure}

In \tabref{tbl:compare_abs_rel_iou_davis}, we compare the average IoU with the average rIoU for the DAVIS and the VOT2016 datasets. Please note that we normalize each tracker with the IoU of the theoretical tracker that has the same abilities. Hence, the KCF tracker is normalized with the {\tt box-no-scale} tracker, the LGT tracker with {\tt box-rot}, and the others with {\tt box-axis-aligned}. By these means, it is possible to observe how well each tracker is doing with respect to its abilities. For the DAVIS dataset, the KCF, ANT, L1APG, and LGT trackers all have the same absolute IoU, but when normalized by $\Phi_{opt}$, differences are visible. Hence, it is evident that the KCF is performing very well, given the fact that it does not estimate the scale. On the other hand, the LGT tracker, which has three more degrees of freedom, is relatively weak. A more detailed example analysis of the {\tt bmx-trees} sequence from DAVIS \cite{perazzi_2016_benchmark} is displayed in \figref{fig:experiments}.

\begin{figure}
\centering
\subfloat[{\tt car1}] {\includegraphics[width=0.215\textwidth]{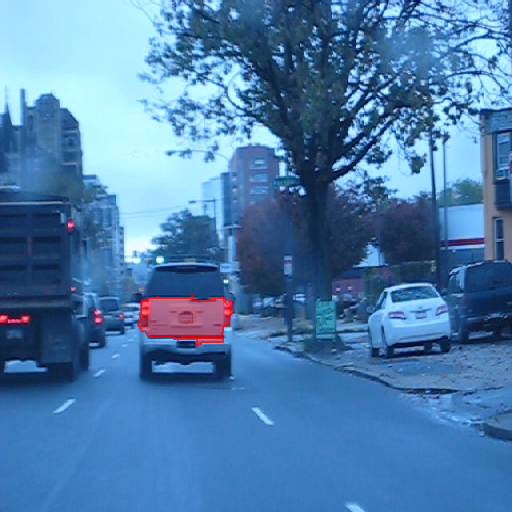}}
  \hfill
\subfloat[{\tt hand}] {\includegraphics[width=0.215\textwidth]{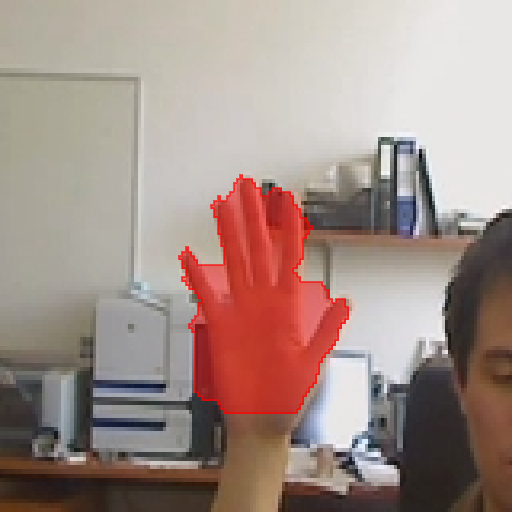}}
  \hfill
\subfloat[{\tt singer2}] {\includegraphics[width=0.215\textwidth]{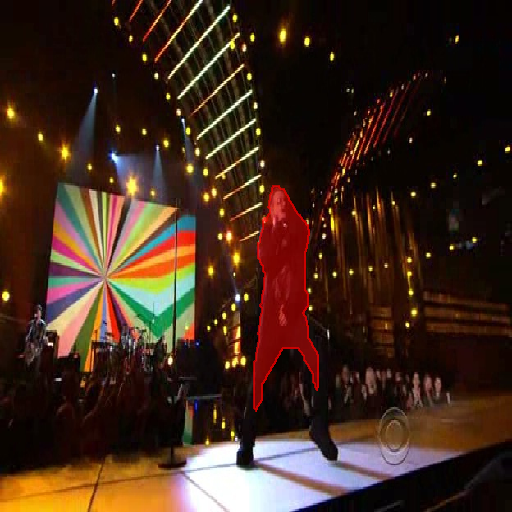}}
  \hfill
\subfloat[{\tt fish3}] {\includegraphics[width=0.215\textwidth]{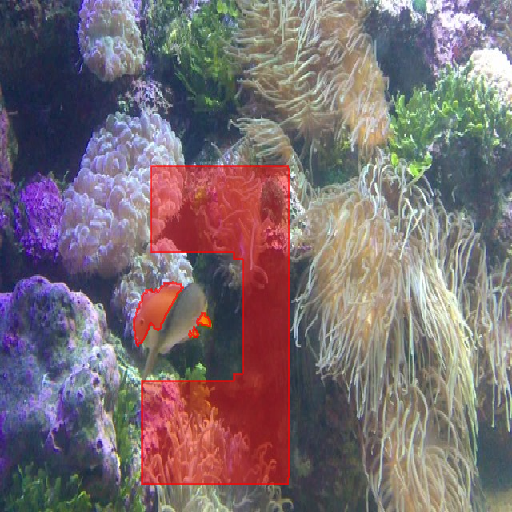}}
  \hfill
\caption{All images are from the VOT2016 \cite{vot_2016} benchmark. Examples where the automatic segmentation used for the VOT 2016 data has difficulties. Either the segmentations are noisy due to motion blur (e.g., (a) and (b)) or there is a weak contrast of the object and its background (c). A handful of scenes have a degenerated segmentation (d).}
\label{fig:votsegmentation_faults}
\end{figure}

For the VOT2016 dataset, the overall accuracies are significantly worse than for DAVIS. On the one hand, this is due to the longer, more difficult sequences, and, on the other hand, due to the less accurate and noisier segmentations (see \figref{fig:votsegmentation_faults}). Nevertheless, the rIoU allows a more reliable comparison of different trackers. For example, ANT, LGT and DSST have almost equal average IoU value, while ANT clearly outperforms LGT and DSST with respect to rIoU. Again, we can see that the KCF tracker is quite strong regarding the fact that it cannot estimate the scale.

\section{Conclusion}
In this paper, we have proposed a new accuracy measure that closes the gap between densely segmented ground truths and box detectors and trackers. We have presented an efficient optimization scheme to obtain the best possible detection boxes for arbitrary segmentations that are required for the new measure. The optimization was validated on a diverse set of segmentations from the DAVIS dataset \cite{perazzi_2016_benchmark}. The new accuracy measure can be used to generate three very expressive theoretical trackers, which can be used to obtain meaningful accuracies and help to interpret scenes without requiring by-frame labels. We have evaluated state-of-the-art trackers with the new accuracy measure on all segmentations within the DAVIS \cite{perazzi_2016_benchmark} and VOT2016 \cite{vot_2016} datasets to display its advantages. The complete code and evaluation system will be made available to the community to encourage its use and make it easy to reproduce our results.

\bibliographystyle{plain}
\bibliography{trackingbib}

\end{document}